\documentclass{article}

\usepackage{arxiv}
\usepackage{natbib}
\usepackage[utf8]{inputenc} % allow utf-8 input
\usepackage[T1]{fontenc}    % use 8-bit T1 fonts
\usepackage{hyperref}       % hyperlinks
\usepackage{url}            % simple URL typesetting
\usepackage{booktabs}       % professional-quality tables
\usepackage{amsfonts}       % blackboard math symbols
\usepackage{nicefrac}       % compact symbols for 1/2, etc.
\usepackage{microtype}      % microtypography
\usepackage{lipsum}
\usepackage{graphicx}
\usepackage{color,soul}
\usepackage{listings}
\usepackage{tabularx}
\usepackage{wrapfig} 
\usepackage{xcolor}

\graphicspath{ {./images/} }

\title{ALKAFI-LLAMA3: Fine-Tuning LLMs for Precise Legal Understanding in Palestine}

\author{
 Rabee Al-Qaesm \\
 AI developer \\
 Faculty of Graduate Studies\\
 Arab American University, Palestine (AAUP)\\
 \texttt{R.alqasem@student.aaup.edu} \\
 \And
 Mohannad Hendi \\
 AI developer \\
 Faculty of Graduate Studies\\
 Arab American University, Palestine (AAUP)\\
 \texttt{m.hendi1@student.aaup.edu} \\
 \And
 Banan Tantour \\
 Legal Advisor \\
 Faculty of Graduate Studies\\
 Birzeit University, Palestine\\
 \texttt{Btantour@birzeit.edu} \\
}

\begin{document}
\maketitle
\begin{abstract}
Large Language Models (LLMs) have demonstrated remarkable potential in diverse domains, yet their application in the legal sector, particularly in low-resource contexts, remains limited. This study addresses the challenges of adapting LLMs to the Palestinian legal domain, where political instability, fragmented legal frameworks, and limited AI resources hinder effective machine-learning applications. We present a fine-tuned model based on a quantized version of Llama-3.2-1B-Instruct, trained on a synthetic data set derived from Palestinian legal texts. Using smaller-scale models and strategically generated question-answer pairs, we achieve a cost-effective, locally sustainable solution that provides accurate and contextually relevant legal guidance. Our experiments demonstrate promising performance on various query types, ranging from yes / no questions and narrative explanations to complex legal differentiations, while highlighting areas for improvement, such as handling calculation-based inquiries and structured list formatting. This work provides a pathway for the deployment of AI-driven legal assistance tools tailored to the needs of resource-constrained environments.

\end{abstract}

% keywords can be removed

\section{Introduction}
Large language models (LLMs) have gained significant attention over the past few years, particularly following the emergence of ChatGPT, both from researchers \cite{movva2024topics} and in terms of their adoption in the private sector. This hype has revolutionized how we use AI in different fields and helped redefine how various domains utilize AI, such as in medicine \cite{alghamdi2024towards,yuan2024continued}, finance \cite{xie2024pixiu,malaysha2024arafinnlp}, and even agriculture \cite{gupta2024rag}. These advancements demonstrate the profound potential of AI to transform industries, driving innovation and efficiency in ways previously unimaginable. However, one domain that still has room for growth and the potential to bring about significant change is the legal domain \cite{martin2024better,maree2024transforming}.

Although sectors such as healthcare and finance have rapidly adopted AI to address their unique challenges, the legal industry has been relatively slow to embrace these technologies \cite{legg2020artificial}. The complex nature of legal language, coupled with jurisdictional variations and the high stakes involved, and the lack of AI regulations \cite{de2021artificial,nadjia2024impact}, has presented significant obstacles to developing effective AI-powered legal solutions. However, the potential of AI to revolutionize legal practice is huge, offering substantial benefits to both legal professionals and citizens \cite{davis2020future}.

\section{Problem statement}

Over the years, the Palestinian legal system has undergone numerous changes due to political instability and prolonged occupation. As a result, legal frameworks lack consistency and clarity, leaving many people struggling to understand their rights and obligations. This situation is further complicated by citizens' general lack of legal knowledge and the absence of accessible, 24/7 legal consultation services. These issues create significant barriers for people seeking timely and reliable legal guidance, often leaving them without the support to navigate their legal concerns effectively.

An AI-powered legal support chatbot can help bridge these gaps by providing accessible assistance around the clock, empowering citizens to address their legal issues confidently and clearly.

\section{Objectives of the Study}
This paper presents a solution for adapting AI technologies, such as large language models (LLMs), to the legal domain. We also explore how such solutions can be developed in countries facing significant AI integration challenges. These challenges include a lack of funding for AI-based projects, limited access to the computational resources required for training such models \cite{omar2024attitudes}, and the absence of AI regulations in these regions \cite{khan2024artificial,melkamu2025artificial}. We demonstrate the feasibility of leveraging smaller models that can be trained and fine-tuned on local machines while utilizing the capabilities of existing LLMs to generate data tailored to their specific problems. This approach facilitates adapting AI applications to process official documents, making them accessible to the public and addressing these countries' unique constraints.

\section{Related Work}

Many studies have explored the complexities of domain-specific training and fine-tuning, mainly aimed at improving the effectiveness of specialized datasets in the legal field. A notable example is SaulLM-7B \cite{colombo2024saullm}, which marks a groundbreaking step forward as a large language model designed to comprehend and generate legal texts. This model is built on the robust Mistral 7B architecture. It harnesses an impressive corpus of more than 30 billion English legal tokens, enabling it to deliver state-of-the-art performance in processing and interpreting complex legal documents. Furthermore, the research introduces an innovative instructional fine-tuning strategy that further bolsters SaulLM-7B’s aptitude for various legal tasks.

In a similar vein, Juru \cite{junior2024juru} concentrates specifically on Brazilian legal texts to minimize the computational costs associated with pre-training. This model utilizes a collection of 1.9 billion unique tokens from credible legal references, illustrating the advantages of domain specialization even when relying on a smaller pool of pretraining data. While this approach can yield effective results within its specific area, it may sacrifice performance in broader knowledge domains.

InternLM-Law \cite{fei2024internlm} takes a targeted approach to tackle the challenges posed by Chinese legal inquiries. The model is trained on an extensive dataset comprising over 1 million legal queries. It employs a two-stage fine-tuning process, beginning with an initial phase incorporating legal and general-purpose content, followed by a focused fine-tuning stage on high-quality legal data. This methodology has achieved state-of-the-art outcomes on the LawBench benchmark, demonstrating superior performance compared to GPT-4 across various subtasks. This highlights the advantages of employing tailored training strategies for specialized legal applications.

\section{Data}

Laws are often enacted to address and regulate various situations that arise to meet human needs. Legal systems can differ significantly from country to country, aligning with the specific requirements of each state's legislation and political orientation. This variation in regulations reflects each country's unique circumstances, priorities, and cultural values \cite{robinson1997politics}. Given the diverse nature of legal systems, it is essential to understand the specificities of each one, especially when creating contextually relevant datasets. Due to the lack of publicly available legal datasets, we created our own, which was vital for effectively fine-tuning our model for accurate legal interpretation within the Palestinian context.

The Palestinian legislative system under the Palestinian Authority comprises several key components that organize state affairs and safeguard citizens' rights. These components work together to implement policies and enforce laws in alignment with the Basic Law. The main elements of this system include:

\begin{itemize}
    \item Constitution: The fundamental legal framework defining the structure of government, rights, and duties in Palestine.
    \item Laws: A set of rules and legislation the government adopts to regulate public and private affairs.
    \item Regulations: Detailed rules issued by relevant authorities to implement and clarify existing laws in Palestine.
    \item Cabinet Decisions: Executive orders made by the Palestinian Cabinet to direct administrative actions and policies.
\end{itemize}

The basic laws are among the key components of the Palestinian legislative system and play a fundamental role in shaping subsequent regulations and bylaws. Therefore, for this study, we concentrated on the basic laws to ensure a foundational understanding of legal regulations.

\subsection{Legal Documents}

In this study, we decided to work with the basic laws adopted within the hierarchy of legislation. These laws are the foundation for regulating legal matters and based on them, the regulations and bylaws that interpret these laws are established and implemented within the Palestinian legislative framework.

We gathered the original law texts used to build the synthetic dataset from the Official Gazette Bureau’s website, the government’s official source for publishing new laws, amendments, government decisions, and other important announcements. We extracted 1,277 text files from this resource, including applicable laws, repealed laws, and amendments to specific laws. We included repealed laws to enlarge the dataset and help the LLM understand Palestine’s legal language, including its terminology, phrasing, and structure, ensuring accurate understanding and interpretation. Following this, we cleaned the text \cite{liu2021faixid,tabassum2020survey} by removing Unicode characters, broken lines of text, repeated characters, multiple new lines, dashes, and any extra spaces before and after the text, though further cleaning steps could still be applied.

\subsection{Synthetic data creation}
Each text file was converted into a structured JSON file to facilitate faster document access. These JSON files contain the article number and corresponding legal text, enabling efficient and organized retrieval.

To create question-and-answer pairs for instruct fine-tuning, we used the ChatGPT API and Gemma API to generate questions and answers based on the legal article text. We designed the prompt, as presented in Table \ref{tab:prompt_example}, to guide the model to generate the pairs in a format that would be easy to integrate with the original JSON files \cite{wu2024eda}. We also used a one-shot learning method \cite{kim2024exploring,nfaoui2024evaluating,isaradech2024zero} to leverage the model to understand the intended output.

\renewcommand{\arraystretch}{1.5} 

\begin{table}[ht]
\centering
\caption{Prompt used for generating question-answer pairs based on legal text.}
\label{tab:prompt_example}
\begin{tabularx}{\textwidth}{|l|X|}
\hline
\textbf{Component}        & \textbf{Description}                                                                                                     \\ \hline
\textbf{Task}             & Write \{NumberOfQuestion\} question(s) and answer(s) from the following legal text.                                       \\ \hline
\textbf{Law Title and Article} & \{LawName\} in Article \{Article Number\}.                                                                         \\ \hline
\textbf{Legal Text}        & \{Legal\_Text\}.                                                                                                        \\ \hline
\textbf{Output Format}     & The output must be in the form of a dictionary: \{question: "", answer: ""\}.                                           \\ \hline
\textbf{Instructions}      & 1. Both the questions and answers must be in Arabic. \newline 2. The question should be written as if someone asked it without legal knowledge. \newline 3. The answer must be written as if provided by a legal advisor. \newline 4. Each answer must begin with the article number and law provided in the text.  \\ \hline
\textbf{Example}           & \textbf{Legal text}: Law No. (4) of 2005 amending Civil Service Law No. (4) of 1998. Amendment to Article (11): Employees in the second category may be promoted to the first category, and employees in the first category may be promoted to the upper category upon meeting the conditions outlined in the law. \newline \textbf{Generated question and answer}: \{ "question": "If I have an employee in the second category, can they be promoted to the first category?", "answer": "Yes, according to Article 2 of Law No. (4) of 2005, which amends Civil Service Law No. (4) of 1998, employees in the second category may be promoted to the first category provided they meet the conditions outlined in the law." \} \\ \hline
\end{tabularx}
\end{table}

This method ensured the consistent generation of question-and-answer pairs tailored to the legal text. It helped establish a robust dataset to support instruct fine-tuning for legal language tasks. Additionally, it streamlined the data generation process and offered a scalable approach for extending similar datasets in future work.

\subsection{Analysis of Dataset Characteristics}

We initially analyzed the question-answer dataset to understand the generated synthetic data better. The dataset contains 243,841 records, each comprising three components: the system message (which includes the legal article and instructions), the user question, and the assistant's answer.

Using the split function, we calculated the total number of words generated, which amounts to approximately 5 million words, with a vocabulary size of 208,835 unique words.

\begin{figure}[htp]
  \centering
  \includegraphics[width=0.75\textwidth]{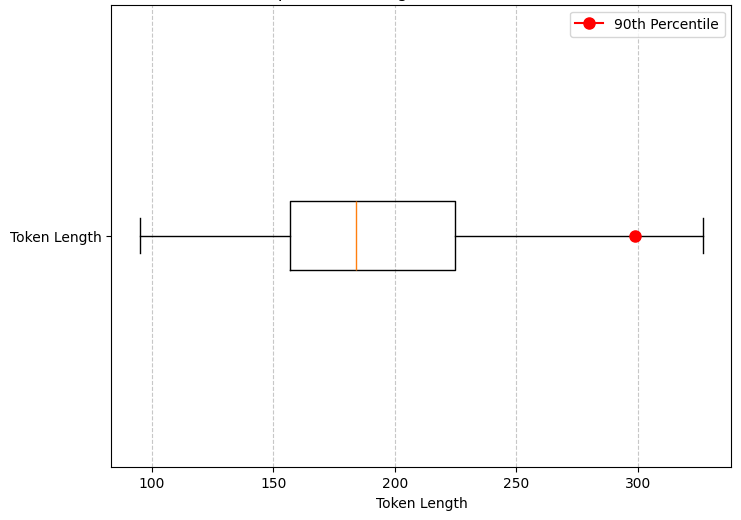}
  \caption{Boxplot of the token length with 90th Percentile.}
  \label{fig:boxplot}
\end{figure}

Next, we analyzed the length of each record for deeper insights into the dataset. The shortest record contains 95 tokens, while the longest spans 2,008 tokens. The median record length is 184 tokens, with 25\% of the records having fewer than 157 tokens and 75\% containing fewer than 225 tokens. On average, the records have a length of approximately 210 tokens. Notably, 90\% of the dataset consists of records with fewer than 299 tokens. This indicates that while most records are concise, a few longer ones significantly exceed the average length, as illustrated in the accompanying graph \ref{fig:boxplot}.

\begin{figure}[htp]
  \centering
  \includegraphics[width=0.75\linewidth]{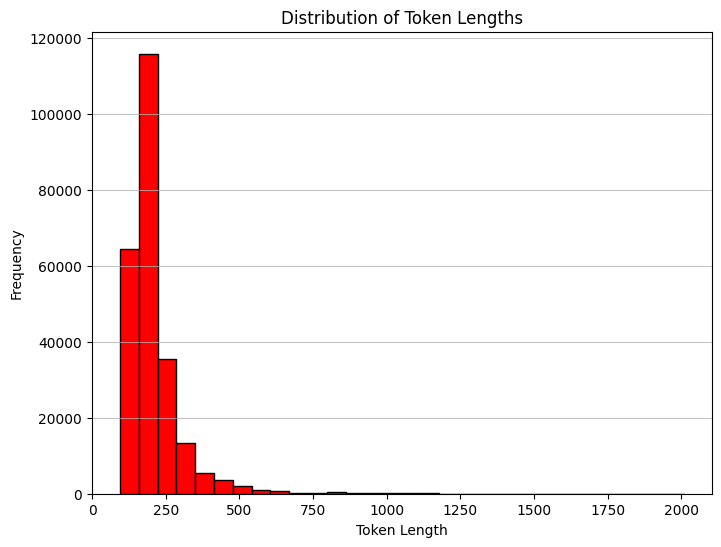}
  \caption{Distribution of the token length across the dataset.}
  \label{fig:dist}
\end{figure}

The distribution of text lengths is heavily skewed towards shorter sentences, which is expected because most legal articles in Palestine are concise. Longer sentences are rare, reflecting the typical structure of legal documents in this context, as shown in the graph \ref{fig:dist}.

Finally, as demonstrated in the graph, we divided the data into training, validation, and testing sets to effectively work with the dataset. This division ensures that the training set contains enough data for model learning while retaining sufficient data in the validation and testing sets for evaluation purposes.

\subsection{Data Availability}
One of the main challenges we faced while conducting this research was the lack of available datasets related to the legal domain, especially in Arab countries. For this reason, we have made the dataset created for this research available for free on Hugging Face \footnote{Dataset is processed and available at Al-Kafi-LR/PalestineLaw-SynthQA}. We hope this dataset can help boost research in the legal field in Arab countries and Palestine.

\newpage
\section{Methodology}

\subsection{Model Selection}

There are numerous open-source LLMs available that can be utilized in our case, ranging from models with one billion parameters, such as those mentioned in \cite{dubey2024llama}, to larger models with up to 180 billion parameters, like Falcon \cite{almazrouei2023falcon}. We decided to use \textbf{Unsloth’s pre-quantized 4bit Llama-3.2-1B-Instruct model [unsloth Llama-3.2-1B-Instruct-bnb-4bit]} for the following reasons:  
\begin{enumerate}
    \item Their strong performance in the Arabic language across various benchmarks \cite{khondaker2024benchmarking}.
    \item Their ability to run efficiently on small GPUs, allows us to operate the model locally on our machines without relying on cloud services.
    \item The quantized version of the model reduces memory usage by approximately 70\%, as demonstrated using Unsloth.
\end{enumerate}
These factors make the Llama models with one billion parameters a compelling choice for our project, as they strike an optimal balance between performance, resource efficiency, and accessibility.

\subsubsection{Fine-Tuning Llama for Palestinian Legal Context}

Even though most generic models \cite{touvron2023llama,jiang2023mistral,achiam2023gpt} rarely encounter Arabic legal data, their responses are often unsatisfactory concerning Palestinian law. They tend to provide answers based on laws from other Arabic countries, making it difficult to deliver accurate responses unless we use the Retrieval-Augmented Generation (RAG) approach. One of the main approaches to enhance these LLMs to have a more accurate answer is to fine-tune \cite{jeong2024fine} these models with domain-specific data, in our case, the Palestinian law.

\subsection{Implementation Details}

\subsubsection{Codebase}
Our code builds upon various open-source frameworks, primarily utilizing Unsloth. This framework leverages Flash Attention from Xformers to optimize transformer-based model training and inference, enabling efficient execution on local machines. It is developed using PyTorch, and all fine-tuned models will be made available on the Hugging Face Hub for accessibility and collaboration.

\subsubsection{Compute}
Our experiments were conducted locally on a system equipped with an \texttt{x86\_64} architecture, featuring an 11th Gen Intel\textregistered{} Core\texttrademark{} i5-11600K processor running at 3.90GHz, 16 GB of RAM, and an NVIDIA GeForce RTX 3060 Ti Hash Rate GPU with 8 GB of VRAM. The operating system used was Ubuntu 22.04.5 LTS. This locally hosted setup provided a robust and efficient environment for training and fine-tuning transformer-based models, enabling the seamless execution of our experiments.

\subsubsection{Enhancing Legal Instruction Adherence}

Since our intention for this model is to use it in the future to support Retrieval-Augmented Generation (RAG) applications for answering legal questions for Palestinian citizens, we decided to fine-tune the Llama instruct model to simulate how RAG works. Specifically, the model is designed to answer questions directly from legal articles provided during the fine-tuning process, mimicking how lawyers respond to people's inquiries based on legal texts.

\subsubsection{Fine-Tuning Arguments}
For the fine-tuning, we ran the model for 10 epochs with a learning rate (LR) of $2 \times 10^{-6}$ with a linear learning rate scheduler and a 10\% warmup ratio to prevent overfitting and complement the existing learned representations in the pre-trained model \cite{wen2024understanding,singh2024warmup}. We employed the Adam optimizer with 8-bit precision for efficiency and LoRA with a rank of 64 for parameter-efficient training. Finally, we set the batch size to one to ensure the model could run without memory problems.

\section{Results}
\begin{figure}[htp]
  \centering
  \includegraphics[width=0.75\linewidth]{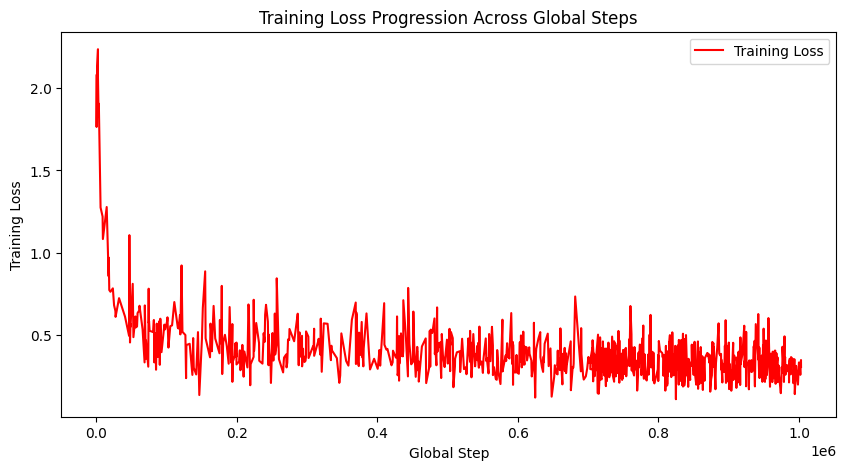}
  \caption{Training loss progression plotted against global steps during the training process.}
  \label{fig:train_loss}
\end{figure}

\begin{figure}[htp]
  \centering
  \includegraphics[width=0.75\linewidth]{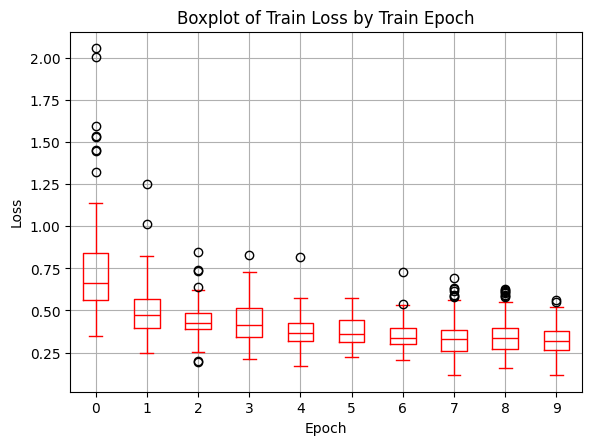}
  \caption{Boxplot illustrating the distribution of training loss across epochs.}
  \label{fig:box_train_loss}
\end{figure}

The fine-tuning process for the model was conducted over 10 days. During this period, the loss function was used to evaluate the performance during the training and evaluation phases. As shown in graph \ref{fig:train_loss} and graph \ref{fig:box_train_loss}, the training loss demonstrates a decrease over the epochs. In the initial stages, the model struggled to learn, resulting in significant spikes in the graph. However, the model stabilized despite minor fluctuations in the later stages, particularly from the second epoch onward. Ultimately, a loss value of 0.33 was achieved.

\newpage

\begin{figure}[htp]
  \centering
  \includegraphics[width=0.75\linewidth]{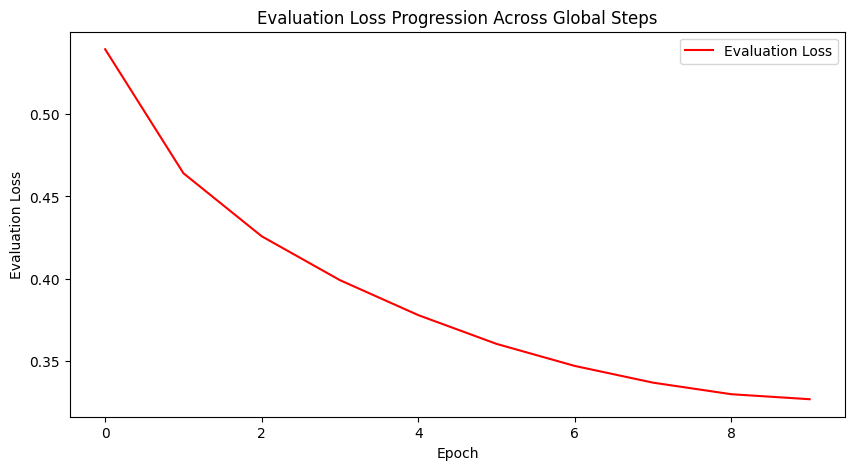}
  \caption{Evaluation loss progression at the end of each epoch.}
  \label{fig:eval_loss}
\end{figure}

For the evaluation, we assessed at the end of each epoch on a small subset of our data to monitor how the model performed on unseen data. The graph \ref{fig:eval_loss} shows that the model's performance improved over time, as evidenced by the decreasing loss, which ultimately reached a value of 0.31.

\section{Experiments}

To experiment with the model's understanding of legal knowledge in the Palestinian context, we asked our domain expert to pose different questions leading to various categories of answers, including Yes/No Answers, Narrative/Explanatory Answers, List-Based Answers, Conditional Answers, and Calculation Answers. This categorization helps ensure a comprehensive analysis of the model's performance across varying scenarios, ranging from narrative explanations to calculations. Each category reflects a unique dimension of legal inquiry.

\subsection*{Yes/No Answers}

\begin{figure}[htp]
  \centering
  \includegraphics[width=\linewidth]{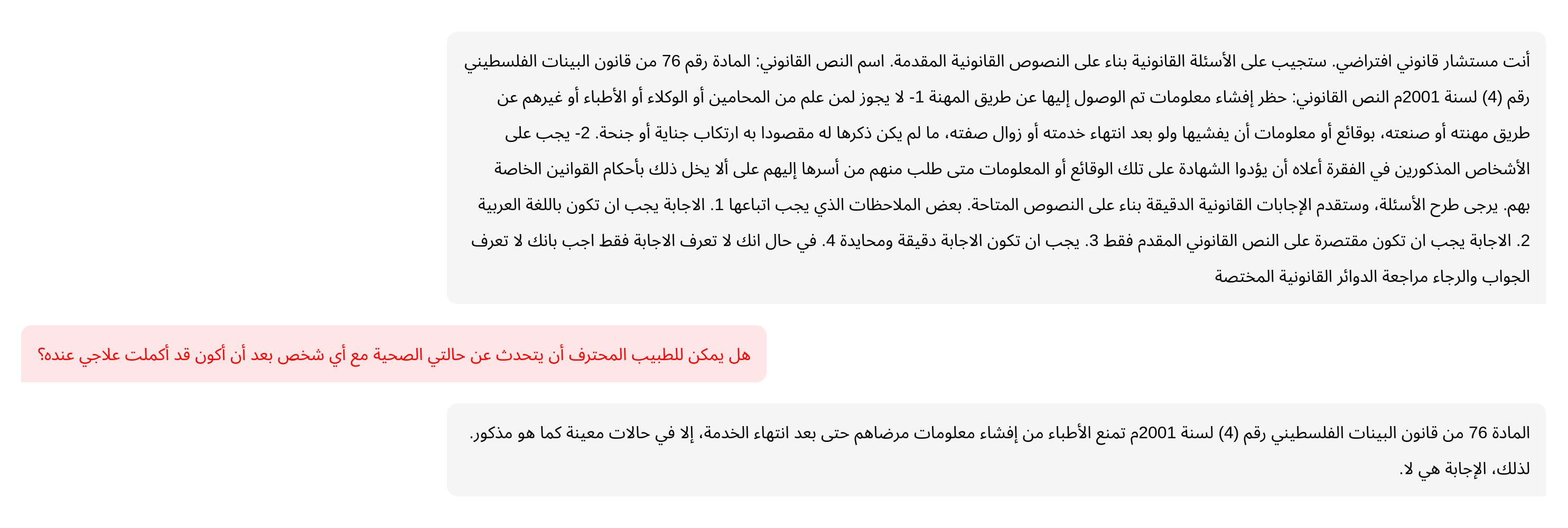}
  \caption{The model's response to a yes/no question}
  \label{fig:yes_no}
\end{figure}

These responses address legal or procedural questions, typically beginning with "Yes" or "No," and often include a brief justification or reference to applicable laws or regulations. For example, in Graph \ref{fig:yes_no}, when we asked the model, "Can a professional doctor discuss my medical condition with anyone after I have completed my treatment with them?" the model, referencing the Palestinian Evidence Law No. 4 of 2001, answered "No." While the response was correct, it can be noted that the model repeated part of the last sentence unnecessarily.

\subsection*{Narrative/Explanatory Answers}

\begin{figure}[htp]
  \centering
  \includegraphics[width=\linewidth]{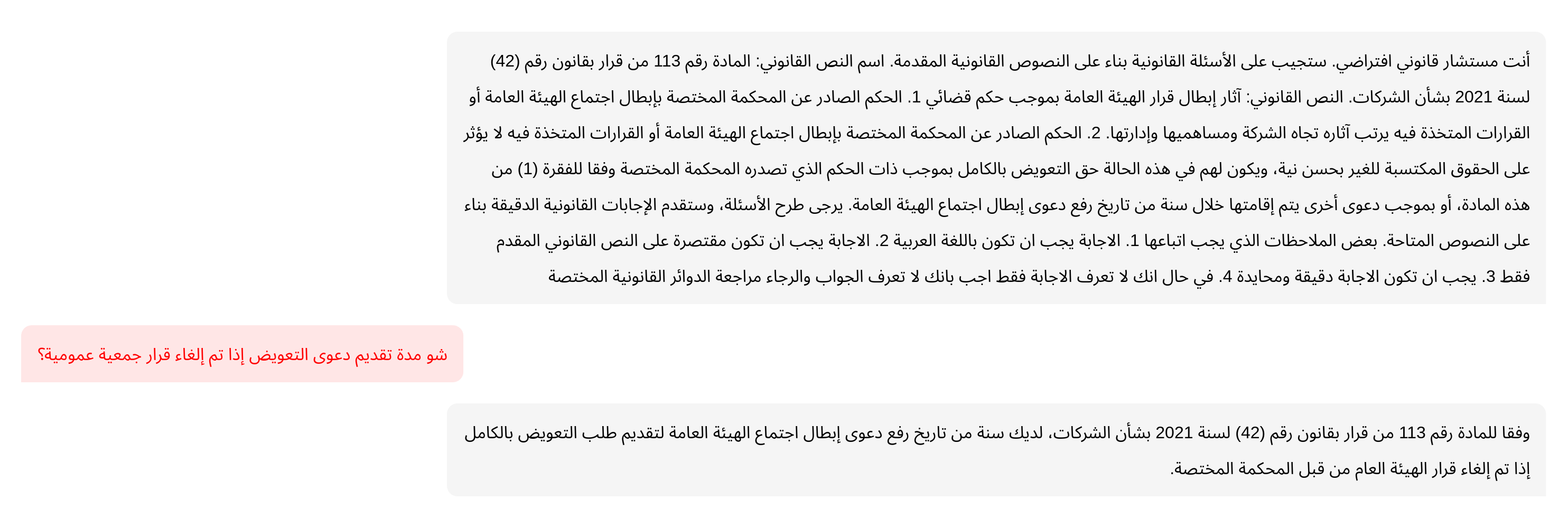}
  \caption{The model's response to a Narrative Question}
  \label{fig:Narr_answers}
\end{figure}

This test evaluated the model's capacity to generate detailed explanations, legal interpretations, or scenario descriptions rather than providing simple yes/no answers or calculations. Graph \ref{fig:Narr_answers} shows the model's response when we asked it about the period in which the model gave the correct answer for the text and mentioned the full article text.

\subsection*{Calculation Answers}
Some articles in various laws include equations to calculate specific matters, such as financial entitlements for workers upon resignation, determining allowances for government employees and even formulas for calculating the number of years of imprisonment for committing a crime. 

In this context, we wanted to test the ability to find the exact result using the equation provided in Article 42 of the Palestinian Labor Law No. (7) of 2000, which outlines financial entitlements upon resignation. This article states that when employees resign, they receive one-third of their annual salary for each year worked if they have been employed for less than five years. 

To test this, we posed the following question: "I am an employee working at Rabee Al-Qasem General Trading Company, and my salary is 5,000 shekels. I have worked for three years. How much are my entitlements?" Unfortunately, the model gave an incorrect answer, applying the wrong equation. This outcome highlights the need for more data on similar situations to help the model understand these questions accurately.

\begin{figure}[htp]
  \centering
  \includegraphics[width=\linewidth]{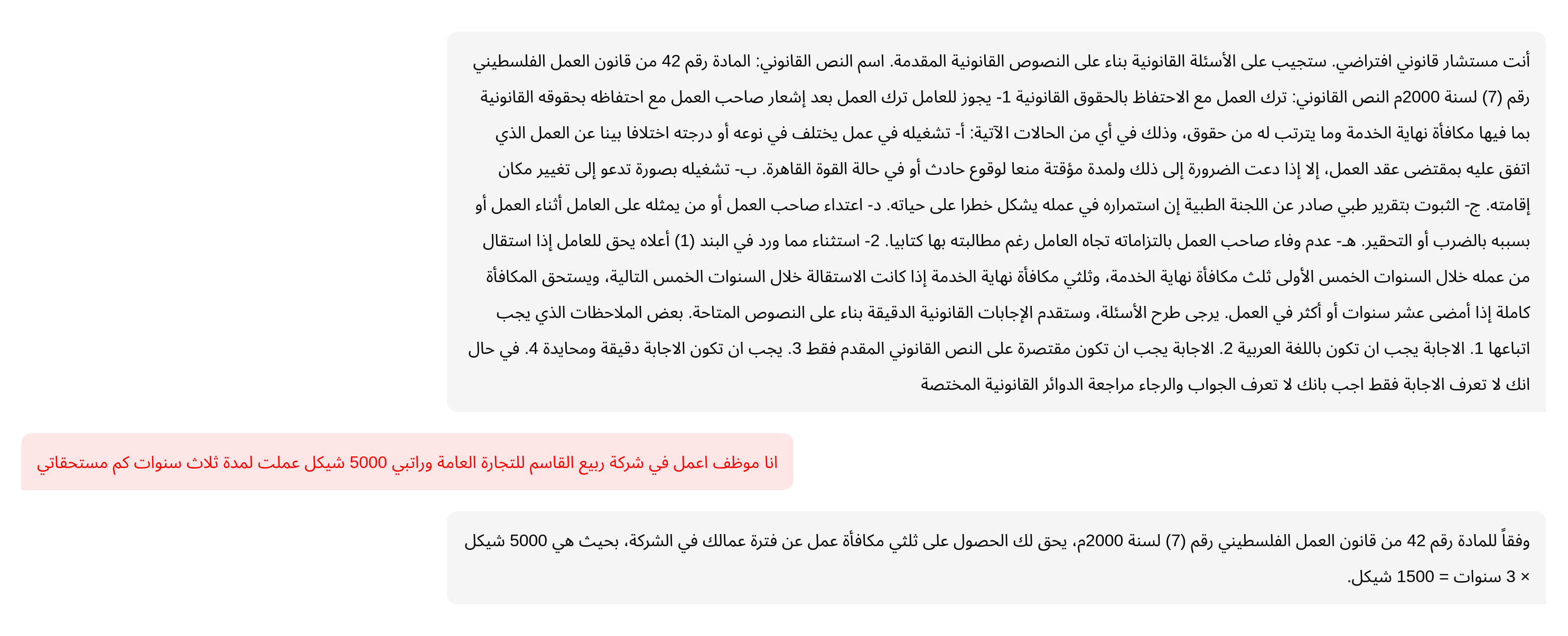}
  \caption{The model's response to a question regarding calculating entitlements from the Palestinian Labor Law.}
  \label{fig:cal}
\end{figure}

\subsection*{List-Based Answers}

Many articles include conditions and enumerated points, such as the conditions required to register a company or a member's qualifications to join a cooperative. We aimed to test the model's ability to answer questions that require listing multiple points, items, or conditions. 

\begin{figure}[htp]
  \centering
  \includegraphics[width=\linewidth]{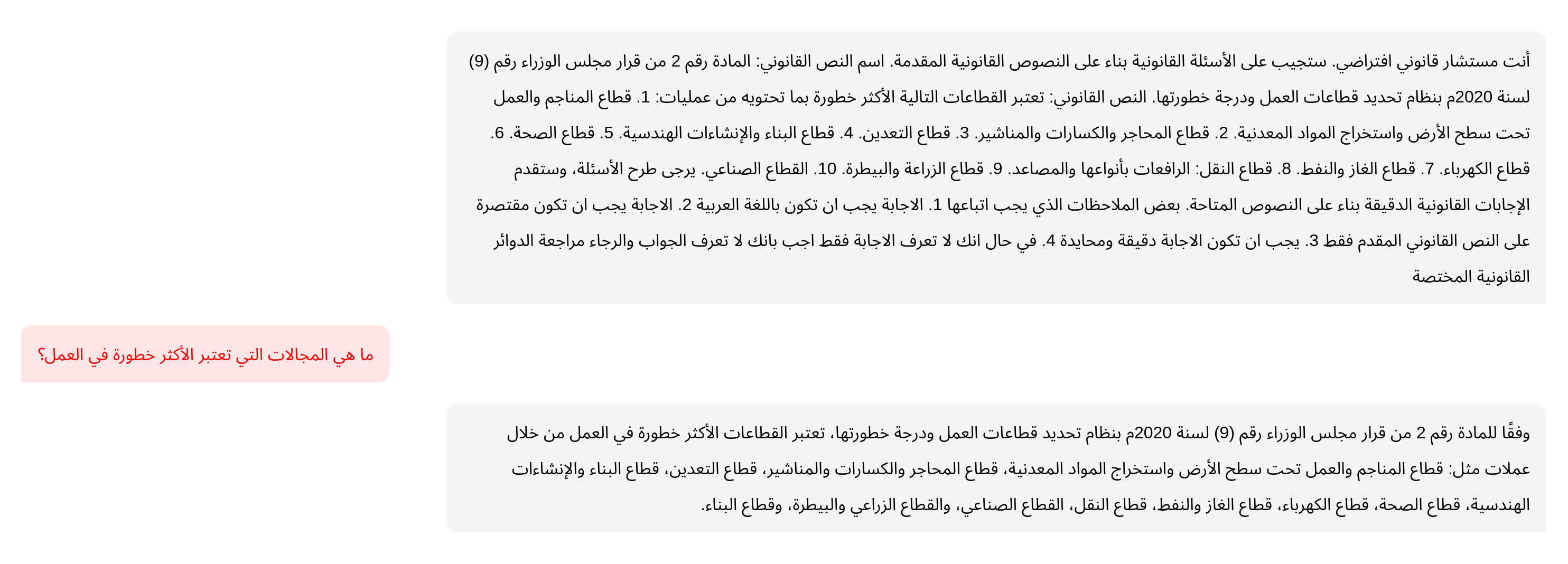}
  \caption{The model's response to a question requiring a list-based answer.}
  \label{fig:list}
\end{figure}

In this test, we asked, "What areas of work are considered the most dangerous?" The expected answer was a list of dangerous areas. While the model was able to identify and provide the relevant areas, it failed to format the response as a proper list. Instead, it presented the items in a single paragraph without numbering or bullet points. This behavior is illustrated in Figure \ref{fig:list}.

\subsection*{Comparative or Differentiation Answers}
Finally, one of the most important aspects of model prediction is to help distinguish between two things and highlight the differences. To test this, we asked the model: "What is the difference between the testimony of one person and that of a group of people?" The model provided a very good, simple, yet clear answer:
"According to Article 1735 of the Ottoman Code of Civil Law, the testimony of a single person is generally valid but may be prone to deception due to individual reasoning. On the other hand, testimony from a group of people is less likely to involve falsehood or deception."

\begin{figure}[htp]
  \centering
  \includegraphics[width=\linewidth]{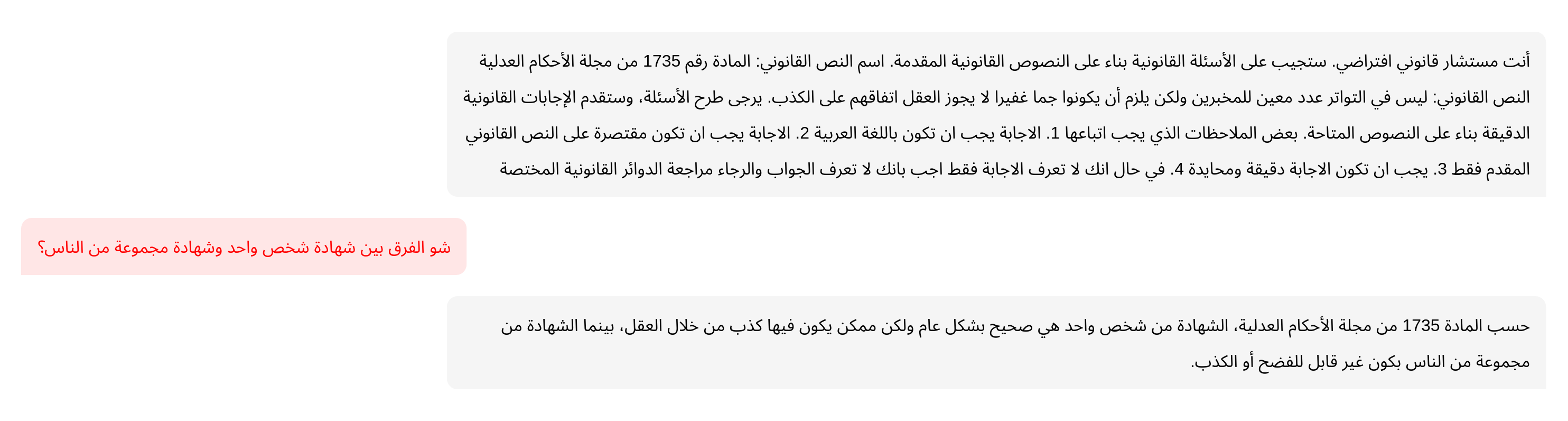}
  \caption{The Model Explains the Difference Between Two Concepts.}
  \label{fig:comp}
\end{figure}

\newpage
\section{Conclusion}

This study presents a pioneering approach to adapting large language models (LLMs) for the Palestinian legal domain. By fine-tuning a quantized version of the Llama-3.2-1B-Instruct model on a synthetic dataset derived from Palestinian legal texts, we demonstrated the feasibility of using resource-efficient models to address the unique challenges of low-resource languages and specialized legal domains. Our results highlight the potential of fine-tuned LLMs to provide accurate legal guidance, overcoming barriers such as limited data availability and computational resources. While the model successfully answered various legal queries, the evaluation also revealed areas where performance could be further enhanced, particularly for complex scenarios like calculations and structured list-based answers.

\bibliographystyle{plainnat}  
\bibliography{references}  %%% Remove comment to use the 

\end{document}